\documentclass[conference]{IEEEtran}
\IEEEoverridecommandlockouts
\usepackage{cite}
\usepackage{amsmath,amssymb,amsfonts}
\usepackage{algorithmic}
\usepackage{graphicx}
\usepackage{textcomp}
\usepackage{xcolor}
\def\BibTeX{{\rm B\kern-.05em{\sc i\kern-.025em b}\kern-.08em
    T\kern-.1667em\lower.7ex\hbox{E}\kern-.125emX}}
\begin{document}

\title{Fine-Tuning LLMs for Reliable Medical Question-Answering Services
}

\author{
    \IEEEauthorblockN{
        Ali Anaissi\IEEEauthorrefmark{1}\IEEEauthorrefmark{2},
        Ali Braytee\IEEEauthorrefmark{2},
        Junaid Akram\IEEEauthorrefmark{1}
        }
    
    \IEEEauthorblockA{
        \IEEEauthorrefmark{1}School of Computer Science, The University of Sydney, Australia \\
        \IEEEauthorrefmark{2}TD School, The University of Technology Sydney, Australia \\
ali.anaissi@sydney.edu.au, ali.braytee@uts.edu.au, junaid.akram@sydney.edu.au
    }}

\maketitle

\begin{abstract}
We present an advanced approach to medical question-answering (QA) services, using fine-tuned Large Language Models (LLMs) to improve the accuracy and reliability of healthcare information. Our study focuses on optimizing models like LLaMA-2 and Mistral, which have shown great promise in delivering precise, reliable medical answers. By leveraging comprehensive datasets, we applied fine-tuning techniques such as rsDoRA+ and ReRAG. rsDoRA+ enhances model performance through a combination of decomposed model weights, varied learning rates for low-rank matrices, and rank stabilization, leading to improved efficiency. ReRAG, which integrates retrieval on demand and question rewriting, further refines the accuracy of the responses. This approach enables healthcare providers to access fast, dependable information, aiding in more efficient decision-making and fostering greater patient trust. Our work highlights the potential of fine-tuned LLMs to significantly improve the quality and accessibility of medical information services, ultimately contributing to better healthcare outcomes for all.
\end{abstract}


\section{Introduction}

The integration of artificial intelligence into healthcare has shown immense potential to improve both the quality and efficiency of medical services \cite{lin2024data,10547221,10492460}. Among the many AI-driven advancements, medical question-answering (QA) systems, powered by Large Language Models (LLMs), have emerged as a key tool for providing accurate and timely answers to complex medical questions \cite{yuan2024large,xiao2023powering,8748975}. This study seeks to push the boundaries of medical QA by utilizing advanced LLMs such as LLaMA-2 and Mistral, which have demonstrated remarkable abilities to understand and generate human-like text. By improving how healthcare professionals access vital information, our work aims to enhance decision-making, promote better patient care, and ensure that accurate medical knowledge is available to all, contributing to the overall well-being of society.

Several fine-tuning techniques have been developed to improve the performance of LLMs. Notable among these are LoRA (Low-Rank Adaptation)\cite{hu2021lora}, rsLoRA (Rank-Stabilized LoRA)\cite{kalajdzievski2023rank}, and DoRA (Weight-Decomposed Low-Rank Adaptation)\cite{liu2024dora}. LoRA facilitates efficient parameter fine-tuning by adding trainable low-rank adapters to specific model layers. RsLoRA enhances this method by introducing a rank stabilization scaling factor, thereby preventing gradient collapse and ensuring stable learning at higher ranks\cite{zhangEnhancingLargeLanguage2024}. DoRA further refines the approach by decomposing the weights of pre-trained models into magnitude and direction matrices, allowing for more precise model adaptation.

Despite these advancements, significant challenges persist in the domain of medical QA\cite{baker2023artificial}. Existing systems frequently encounter difficulties in comprehending complex medical terminology and delivering contextually accurate responses\cite{zhangSirenSongAI2023a,8617007}. Additionally, the continuously evolving nature of medical knowledge necessitates frequent updates, which are often computationally intensive and time-consuming. Moreover, current models are susceptible to inaccuracies and may propagate outdated or incorrect information, thereby compromising their reliability and trustworthiness in critical healthcare scenarios \cite{choudhury2024large, mesko2023imperative}.

To address these challenges, we propose a novel approach that synergizes advanced fine-tuning techniques with an innovative retrieval-augmented generation method. We introduce rsDoRA+, which combines the benefits of rsLoRA and DoRA with different learning rates for low-rank matrices. Additionally, we propose a new method, ReRAG (Retrieval on Demand and Question Rewrite), which integrates retrieval and question rewrite components to further enhance model performance. RsDoRA+ improves the model's understanding of medical terminology, reasoning, and contextual accuracy by leveraging decomposed model weights with rank stabilization. ReRAG optimizes the model's performance by providing relevant information on demand and ensuring the retrieval of the most pertinent data.

Our contributions in this study are as follows:
\begin{itemize}
    \item Introduction of rsDoRA+, a fine-tuning technique that combines decomposed model weights with rank stabilization and differential learning rates for low-rank matrices, thereby significantly enhancing the performance of LLMs in medical QA services.
    \item Development of ReRAG, a retrieval-augmented generation method that integrates on-demand retrieval and question rewrite components to improve the accuracy and relevance of medical QA responses.
    \item Creation of a specialized LLM tailored for medical information retrieval, enabling healthcare providers to deliver fast and reliable responses, thus enhancing decision-making efficiency and patient trust.
    \item Demonstration of the potential for advanced information technology to significantly improve the quality of medical information services through comprehensive experimentation and evaluation.
\end{itemize}

%The structure of this paper is organized as follows. Section 2 provides a comprehensive review of the literature on LLM fine-tuning techniques and their application in medical QA services. Section 3 describes our methodology, including the datasets utilized, fine-tuning techniques employed, and the architecture of our proposed system. Section 4 presents the experimental setup and results, highlighting the improvements achieved by our approach. Finally, Section 5 discusses the implications of our findings, addresses the limitations of our study, and outlines future research directions.

\section{Related Work}

Retrieval-Augmented Generation (RAG) leverages both parametric and non-parametric memory, significantly enhancing the performance of Large Language Models (LLMs) in translation and question-answering tasks, as highlighted by Lewis et al. \cite{lewisRetrievalAugmentedGenerationKnowledgeIntensive2021}. These models, equipped with extensive training datasets and substantial parameters, have led advancements in these fields. However, they face persistent challenges such as the potential for outdated or incorrect information and difficulties with real-time updates \cite{zhangSirenSongAI2023a, zhangEnhancingLargeLanguage2024,hayouLoRAEfficientLow2024, kalajdzievski2023rank}. The RAG approach improves LLM performance through pre-training, combining various memory types to generate fact-based, varied, and accurate language representations. This method employs a dynamic updating mechanism to refresh the knowledge base without retraining the entire model, thereby enhancing reliability and clarity \cite{gaoRetrievalAugmentedGenerationLarge2024}. Nonetheless, RAG faces issues like noise or conflicting information during the retrieval phase, necessitating improvements for response accuracy and reliability \cite{yuAugmentationAdaptedRetrieverImproves2023}. Lin et al. \cite{liuWhenMOEMeets2024} suggest integrating RAG with fine-tuning methods to maximize benefits from both parametric and non-parametric approaches.

SELF-RAG further advances traditional RAG by incorporating selective retrieval and self-reflection mechanisms, thus enhancing the quality and accuracy of language models. Unlike traditional RAG, which may retrieve irrelevant information, SELF-RAG ensures that only relevant content is retrieved based on the model's self-evaluation. It also incorporates self-reflection tokens that allow the model to assess the quality and integrity of its responses, thereby increasing the connectedness and correctness of its outputs \cite{asaiSelfRAGLearningRetrieve2023}. Fine-tuning adjusts the model's weights according to new data, allowing modifications without the need for retraining the entire model. This method is particularly effective in customizing pre-trained LLMs for specific tasks using labeled data, as seen in Supervised Fine-Tuning (SFT) and Parameter-Efficient Fine-Tuning (PEFT) \cite{dongHowAbilitiesLarge2024, zhangEnhancingLargeLanguage2024}. Among PEFT methods, Low-Rank Adaptation (LoRA) and its advanced versions, such as LoRA+ and DoRA, have shown significant improvements. LoRA introduces trainable low-rank matrices within the Transformer architecture, enhancing the efficiency of parameter updates without altering the model architecture significantly \cite{hanParameterEfficientFineTuningLarge2024, hayouLoRAEfficientLow2024}. DoRA further refines this approach by decomposing weights into magnitude and direction matrices, thereby improving robustness and latency \cite{liu2024dora}. Rank-Stabilized LoRA (rsLoRA) addresses limitations in higher-rank learning by scaling adapters by the square root of the rank, thus enabling fine-tuning efficiency and performance without increasing computational costs \cite{kalajdzievski2023rank}. The success of rsLoRA paves the way for future research into more efficient fine-tuning techniques, such as AdaLoRA \cite{zhangAdaLoRAAdaptiveBudget2023}. Overall, these methodologies collectively enable LLMs to handle dynamic, context-sensitive medical information, significantly improving the accuracy and depth of medical responses. The integration of these advanced techniques into LLMs represents a considerable advancement in addressing the challenges of modern healthcare information systems, transforming the decision-making processes among healthcare practitioners.

\section{Proposed Method/Framework}

To enable large language models (LLMs) to adapt efficiently to new data distributions under limited resources, parameter-efficient fine-tuning techniques are essential. Low-Rank Adaptation (LoRA) has emerged as a mainstream method by adding trainable low-rank adapters to selected layers, which are products of low-rank matrices scaled by a factor related to the rank. However, traditional LoRA's aggressive scaling factor limits its performance at higher ranks, leading to slower learning. To address this, Rank-Stable LoRA (rsLoRA) introduces a more conservative scaling factor, enhancing learning stability and performance at higher ranks. Experimental results indicate that rsLoRA significantly improves stability and learning efficiency when using higher ranks.

\begin{figure*}[t]
\centering
\includegraphics[width=0.8\textwidth]{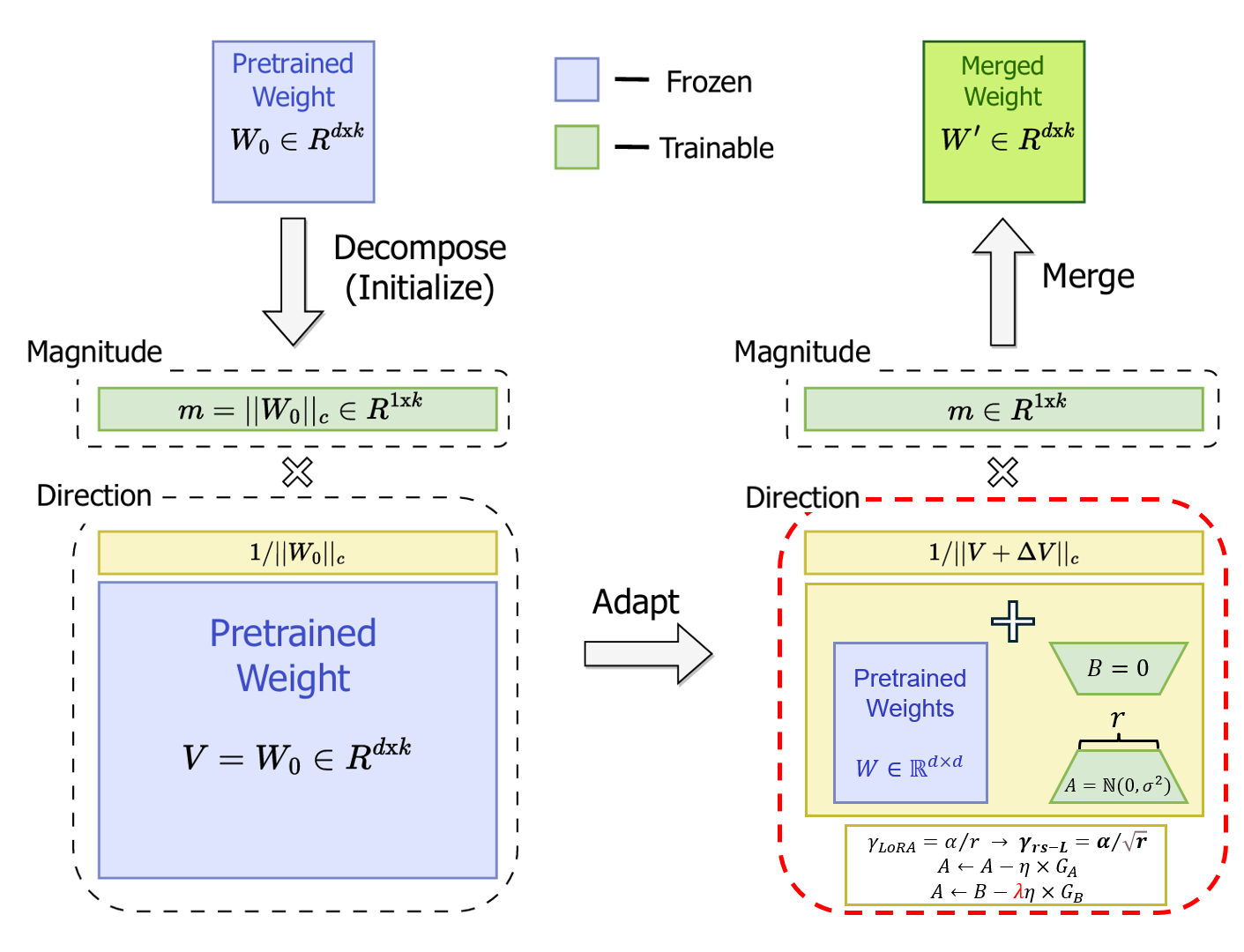}
\caption{Reparameterization of rsDoRA+.}
\label{fig:repara}
\end{figure*}

Moreover, LoRA and rsLoRA show a positive correlation between the magnitude and direction of updates, often proportional, which might limit precise adjustments to model weights. Decomposed Low-Rank Adaptation (DoRA) addresses this by decomposing pre-trained weights into magnitude and direction matrices for fine-tuning, specifically using LoRA for direction updates. This minimizes the number of trainable parameters while enhancing learning capability and stability without additional inference overhead. Additionally, the standard LoRA method uses the same learning rate for adapter matrices $\boldsymbol{A}$ and $\boldsymbol{B}$, leading to inefficiencies in models with high embedding dimensions. LoRA Plus (LoRA+) sets different learning rates for matrices $\boldsymbol{A}$ and $\boldsymbol{B}$, significantly increasing the learning rate of matrix $\boldsymbol{B}$ compared to $\boldsymbol{A}$, improving training efficiency and feature learning ability.

In Natural Language Processing (NLP), adversarial training enhances model robustness by introducing adversarial examples, slight perturbations added to the dataset that are almost imperceptible to humans but can cause the model to make incorrect predictions. This makes the model more resistant to such examples and improves robustness. Adding noise to embeddings during training reduces overfitting to specific details like format, exact wording, and text length of the fine-tuning dataset. Therefore, we added a certain proportion of noise to the embeddings at the LLM base layer to mitigate overfitting during the instruction fine-tuning phase and better utilize the knowledge from the pre-training phase.

Our approach to refining the training process for LLMs, particularly in the embedding layer, involves a strategy called outlining, which enhances the model's ability to adapt and learn from new data efficiently. This process adjusts the learning rates for adapter matrices $\boldsymbol{A}$ and $\boldsymbol{B}$ strategically. Specifically, the learning rate for matrix $\boldsymbol{B}$ is set as a multiple, denoted as $\lambda$, of the learning rate for matrix $\boldsymbol{A}$. This ensures intensive fine-tuning of certain model aspects without overwhelming computational demands, maintaining a balance that allows for efficient and effective learning. This adjustment optimizes learning while preserving the model’s stability and reducing overfitting risks, thereby refining the adaptability of models to manage diverse and complex data scenarios more robustly.

\begin{figure*}[t]
\centering
\includegraphics[width=0.8\textwidth]{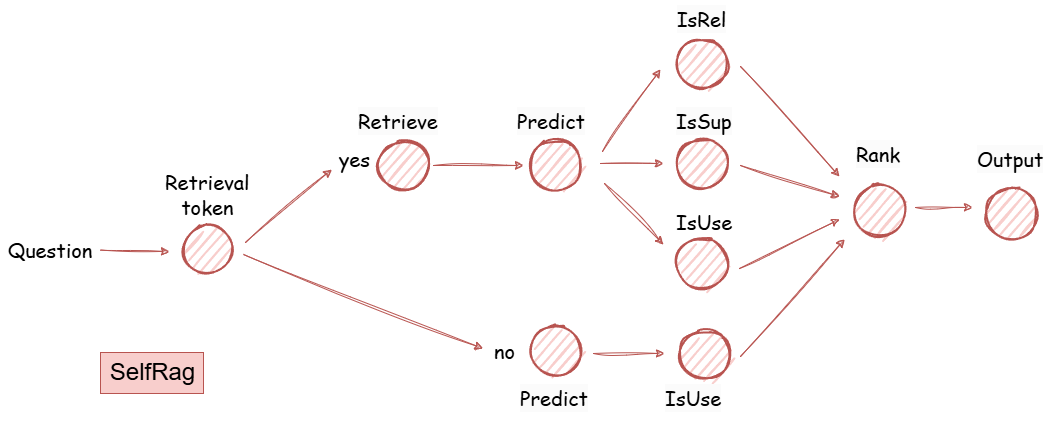}
\caption{Structure of SelfRAG method.}
\label{fig:Structure_SelfRAG}
\end{figure*}

\begin{figure*}[t]
\centering
\includegraphics[width=0.8\textwidth]{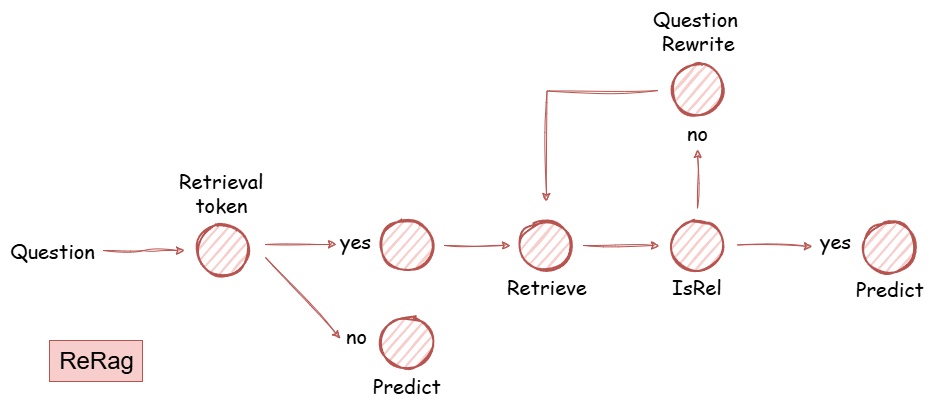}
\caption{Structure of ReRAG method.}
\label{fig:Structure_ReRAG}
\end{figure*}

The ReRAG model, inspired by the selfRAG method (see Figure \ref{fig:Structure_SelfRAG}), introduces tokens to evaluate the necessity and relevance of retrieved text for answering a question. The base RAG model sometimes adds irrelevant or unnecessary text, degrading performance. To enhance this, ReRAG incorporates retrieval tokens and relevance tokens to filter out useless retrieved text and questions that do not require additional retrieval. The simplified structure of ReRAG, depicted in Figure \ref{fig:Structure_ReRAG}, aims to improve performance by integrating a question rewrite component that ensures retrieval of the most useful text.

\begin{table}[t]
\centering \caption{Model Performance and Characteristics}
\begin{tabular}{cccc}
\hline
Model Name & Average Performance & Speed & Model Size \\
\hline
all-MiniLM-L12-v2 & 59.76 & 7500 & 120 MB \\
all-MiniLM-L6-v2 & 58.80 & 14200 & 80 MB \\
\hline
\end{tabular}

\label{tab:embedding}
\end{table}

Initially, the model accepts a question as input, with a critic model determining the retrieval token to decide if retrieval text is necessary. If supplementary information is needed, the question is passed to the retrieval node. The retriever converts the query into a vector embedding using a sentence transformer and calculates the best-suited texts compared to the vector base. The IsRel node then assesses the relevance of each text to the question, discarding irrelevant texts and saving relevant ones for further use. If no relevant text is found, the question rewrite component extracts key phrases from the question to perform a new retrieval, ensuring the most pertinent information is utilized.

For embedding models, a comparison between all-MiniLM-L6-v2 and all-MiniLM-L12-v2 was conducted, as shown in Table \ref{tab:embedding}. The cosine similarity between the question vector and phrase vectors is calculated to select the most similar candidates. Min-max normalization is applied to ensure proportional feature contribution to the Euclidean distance, and the candidates are ordered by their distance sums to enhance phrase diversity.

\begin{equation}
\operatorname{Cosine}(\vec{v}, \vec{w})=\frac{\vec{v} \cdot \vec{w}}{|\vec{v}||\vec{w}|}=\frac{\sum_{i=1}^N v_i w_i}{\sqrt{\sum_{i=1}^N v_i^2} \sqrt{\sum_{i=1}^N w_i^2}}
\label{eq:sim}
\end{equation}

\begin{equation}
x^{\prime}=a+\frac{(x-\min (x))(b-a)}{\max (x)-\min (x)}
\label{eq:norm}
\end{equation}

\begin{equation}
\operatorname{dist}\left(\boldsymbol{X}_1, \boldsymbol{X}_2\right)=\sqrt{\sum_{i=1}^n\left(x_{1 i}-x_{2 i}\right)^2}
\label{eq:distance}
\end{equation}

\begin{equation}
\operatorname{Sum}(\boldsymbol{X}_i) = \sum_{j=1}^n(\operatorname{dist}\left(\boldsymbol{X}_i, \boldsymbol{X}_j\right)
\label{eq:sum}
\end{equation}

Finally, if relevant texts are found in the IsRel node, they are combined with the original question to form a new prompt for the generator model. If no relevant text is found after using the question rewrite component, the generator model uses the original question to generate the final answer. This comprehensive approach ensures that the most relevant information is used to enhance the accuracy and relevance of responses in medical question-answering services.

\section{Results and Discussion}

\subsection{Dataset and Environment Setup}
The question-answering dataset employs data from Medical Meadow, integrating Anki Flashcards and MediQA to showcase our model's generalizability across medical datasets. Medical Meadow, a comprehensive collection sourced from the medAlpaca project and established medical databases, provides diverse medical data. Anki Flashcards, created by medical students, cover various medical disciplines with concise summaries and mnemonics. We used OpenAI's GPT-3.5-turbo to rephrase these flashcards into coherent question-answer pairs. MediQA comprises manually generated summaries of answers to consumer health questions, ensuring rich, relevant data for training.

For rsDoRA+ deployment, we set up the environment by installing necessary libraries, dependencies, and downloading base models for fine-tuning. We automated training and evaluation processes with shell commands, and analyzed metrics such as accuracy, loss, Rouge, and BLEU scores to evaluate model performance. For ReRAG, we installed libraries, set up retrieval and vector databases, and used rsDoRA+-fine-tuned models as generators. Detailed steps are in ReadmeReRAG.md.

\subsection{Experimental Setup and Validation}
To validate rsDoRA+ and ReRAG, we designed experiments using medical dialogue datasets like MediQA and Anki Flashcards. We used 7B models like LLama2 and Mistral for their balance of performance and speed, comparing Full-Parameter Fine-Tuning and LoRA Fine-Tuning baselines with various configurations of rsDoRA+, rsLoRA combined with DoRA, and LoRA+. In ReRAG experiments, using langchain, the base model (LLama2 chat model fine-tuned with rsDoRA+) was tested. Unlike SelfRAG's extensive data generation with GPT-4, we used a resource-efficient GPT-3.5 API approach. The Retrieve Token Node answered with a binary score to decide if detailed information was needed. The Retrieve Node embedded 100-word text pieces using the ‘tao-8k’ model, with the retriever finding the best-suited texts via Euclidean distance. The IsRel Node assessed relevance, discarding irrelevant texts, while the Question Rewrite Node performed new retrievals if needed. Predict Nodes used either the original question or a new query with retrieved texts.

\subsection{Performance Analysis}
\begin{figure}[t]
    \centering
    \includegraphics[width=1\linewidth]{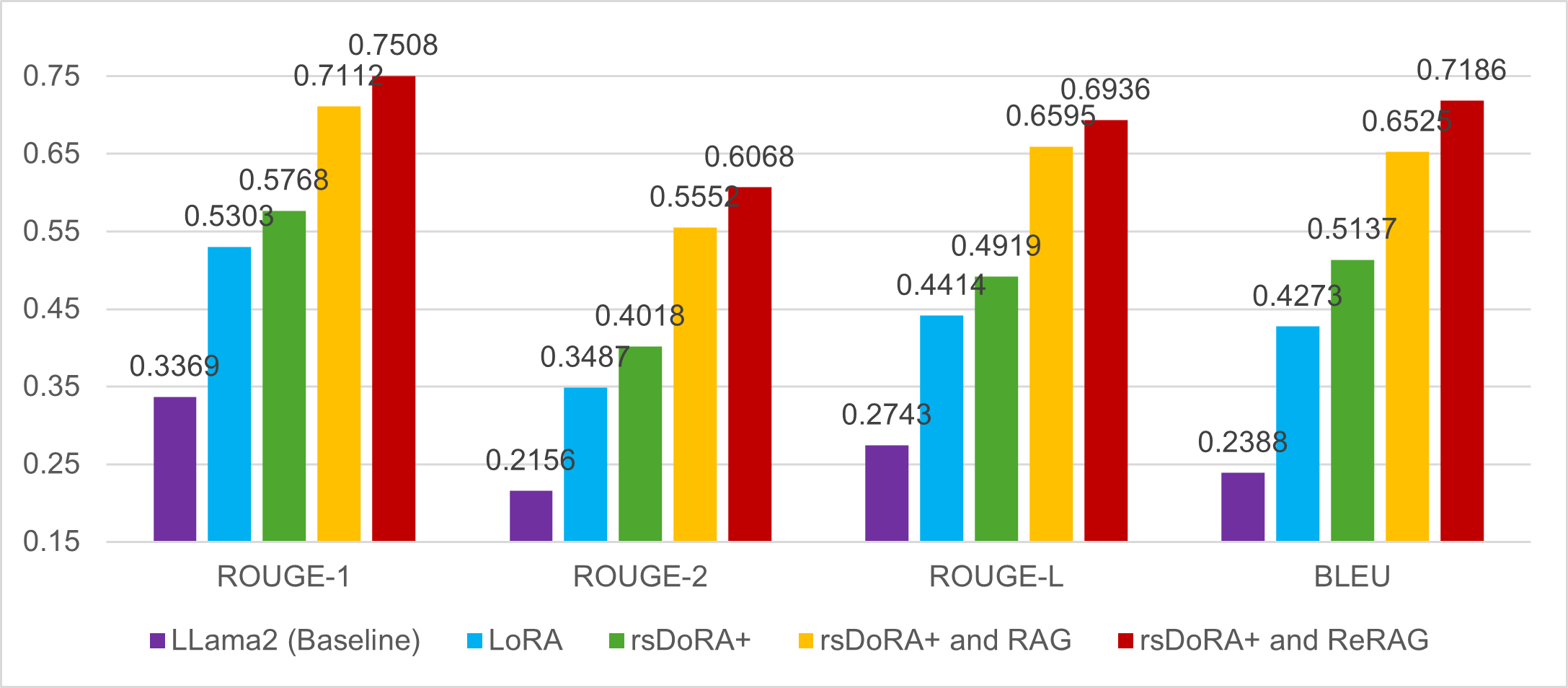}
    \caption{Performance of LLaMA2 with Different $\alpha$ on the Flashcards}
    \label{fig:overallresults}
\end{figure}

For evaluation, we used an independent test set and standard metrics like BLEU and ROUGE to assess performance, ensuring the accuracy and relevance of responses in medical QA services. Our results demonstrate that rsDoRA+ significantly enhances LLaMA2 model performance, while ReRAG improves LLaMA2 performance as shown in Figure \ref{fig:overallresults}. The integration of rsDoRA+ into ReRAG has markedly boosted QA performance in the medical field, with improvements across four metrics being 123\%, 181\%, 153\%, and 201\%, respectively, indicating exceptional performance at high ranks with enhanced QA and answer richness.

rsDoRA+ is a novel fine-tuning method optimized from models like DoRA, combining the advantages of DoRA, LoRA+, and rsLoRA, and incorporating NEFtune to prevent overfitting. Before fine-tuning, rsDoRA+ introduces a certain proportion of noise into the embeddings, sampled from a uniform distribution \([-1, 1]\) and scaled by \(\alpha / Ld\), as shown in Equation \ref{eq:Noise}.

\begin{equation}
\operatorname{Noise} = \frac{\alpha}{L \cdot d} U(-1,1)
\label{eq:Noise}
\end{equation}

\begin{figure}
    \centering
    \includegraphics[width=1\linewidth]{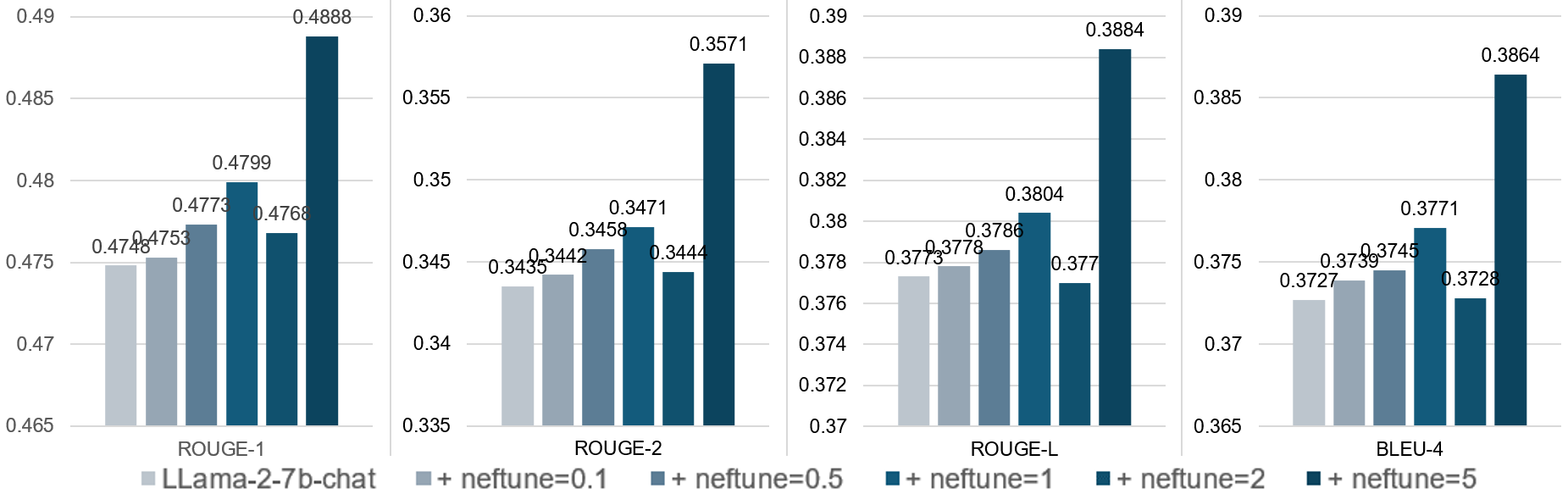}
    \caption{Performance of LLaMA2 with Different $\alpha$ on the MediQA}
    \label{fig:alphaLLaMA2MediQA}
\end{figure}

\begin{figure}[t]
    \centering
    \includegraphics[width=1\linewidth]{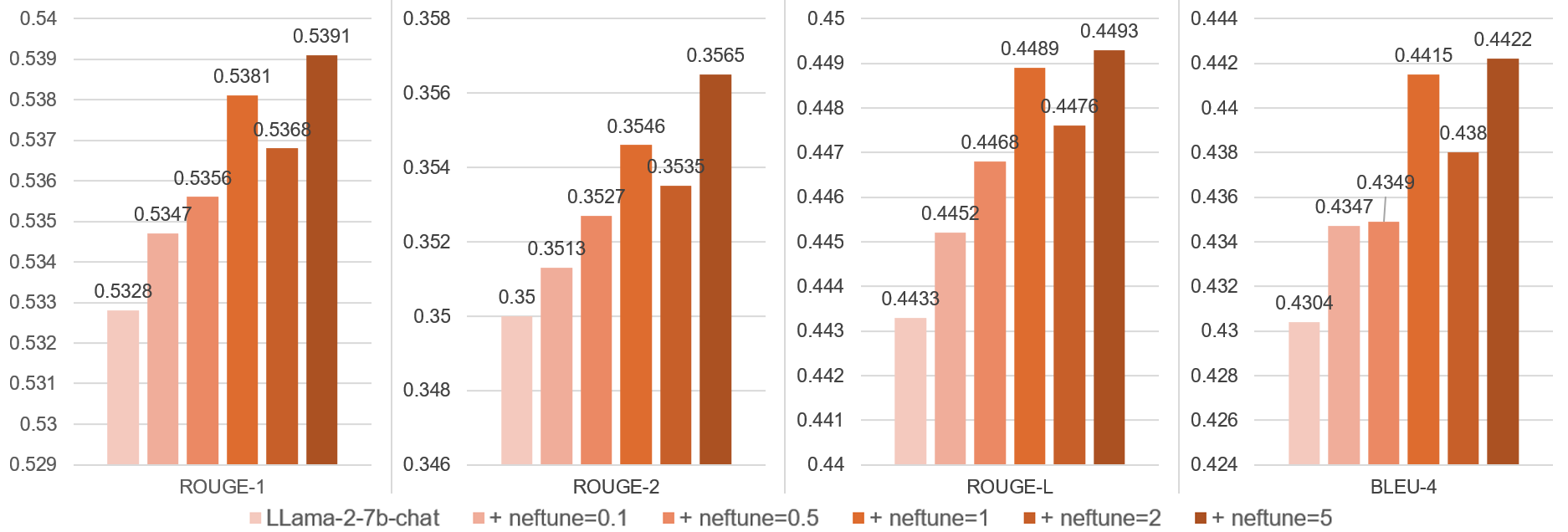}
    \caption{Performance of LLaMA2 with Different $\alpha$ on the Flashcards}
    \label{fig:alphaFlashcards}
\end{figure}

Where $\alpha$ is an adjustable parameter, $L$ is the input length, and $d$ is the embedding dimension. This noise helps mitigate overfitting and better utilize pre-training knowledge. Experiments comparing different values of $\alpha$ using NEFtune were conducted on the MediQA and Anki Flashcards datasets, using LLaMA2-7b-chat as the baseline model with DoRA and rsLoRA fine-tuning. We tested $\alpha$ values of 0.1, 0.5, 1, 2, and 5, as illustrated in Figures \ref{fig:alphaLLaMA2MediQA} and \ref{fig:alphaFlashcards}. NEFtune significantly improved the model's question-answering performance.

\begin{table*}[t]
\centering \caption{Performance of LLaMA2 with Different $\alpha$ on the MediQA}
\begin{tabular}{ccccc}
\hline
Model & ROUGE-1 & ROUGE-2 & ROUGE-L & BLEU-4 \\
\hline
Base & 0.4748 & 0.3435 & 0.3773 & 0.3727 \\
\texttt{+ NEFtune ($\alpha = 0.1$)} & 0.4753 & 0.3442 & 0.3778 & 0.3739 \\
\texttt{+ NEFtune ($\alpha = 0.5$)} & 0.4773 & 0.3458 & 0.3786 & 0.3745 \\
\texttt{+ NEFtune ($\alpha = 1$)} & 0.4799 & 0.3471 & 0.3804 & 0.3771 \\
\texttt{+ NEFtune ($\alpha = 2$)} & 0.4768 & 0.3444 & 0.3770 & 0.3728 \\
\texttt{+ NEFtune ($\alpha = 5$)} & 0.4888 & 0.3571 & 0.3884 & 0.3864 \\
\hline
\end{tabular}
\end{table*}

In the MediQA dataset, the ROUGE-L score was only 0.3\% lower than the baseline when $\alpha = 2$. The most significant improvement was at $\alpha = 5$, showing a consistent performance increase of approximately 2\%. NEFtune's impact varied; it had a smaller effect on individual word overlap but a more substantial impact on the overlap of four-word sequences and longer texts. However, performance did not consistently improve with increasing $\alpha$—at $\alpha = 2$, performance decreased slightly compared to $\alpha = 0.5$ and $\alpha = 1$. This variation may be due to differing hyperparameters, such as the number of fine-tuning epochs.

\begin{table*}[t]
\centering \caption{Performance of LLaMA2 with Different $\alpha$ on the Flashcards}
\begin{tabular}{ccccc}
\hline
Model & ROUGE-1 & ROUGE-2 & ROUGE-L & BLEU-4 \\
\hline
Base & 0.5328 & 0.35 & 0.4433 & 0.4304 \\
\texttt{+ NEFtune ($\alpha = 0.1$)} & 0.5347 & 0.3513 & 0.4452 & 0.4347 \\
\texttt{+ NEFtune ($\alpha = 0.5$)} & 0.5356 & 0.3527 & 0.4468 & 0.4349 \\
\texttt{+ NEFtune ($\alpha = 1$)} & 0.5381 & 0.3546 & 0.4489 & 0.4415 \\
\texttt{+ NEFtune ($\alpha = 2$)} & 0.5368 & 0.3535 & 0.4476 & 0.4380 \\
\texttt{+ NEFtune ($\alpha = 5$)} & 0.5391 & 0.3565 & 0.4493 & 0.4422 \\
\hline
\end{tabular}
\end{table*}

After introducing noise with $\alpha = 5$, we tested the fine-tuning approach and evaluated performance using the BLEU-4 metric. Given that LoRA enables large language models to adapt efficiently to new data distributions, our model was optimized based on the LoRA framework. However, we found that traditional LoRA's aggressive scaling factor slowed the learning speed of high-rank adapters, limiting performance.

\begin{figure}[t]
    \centering
    \includegraphics[width=1\linewidth]{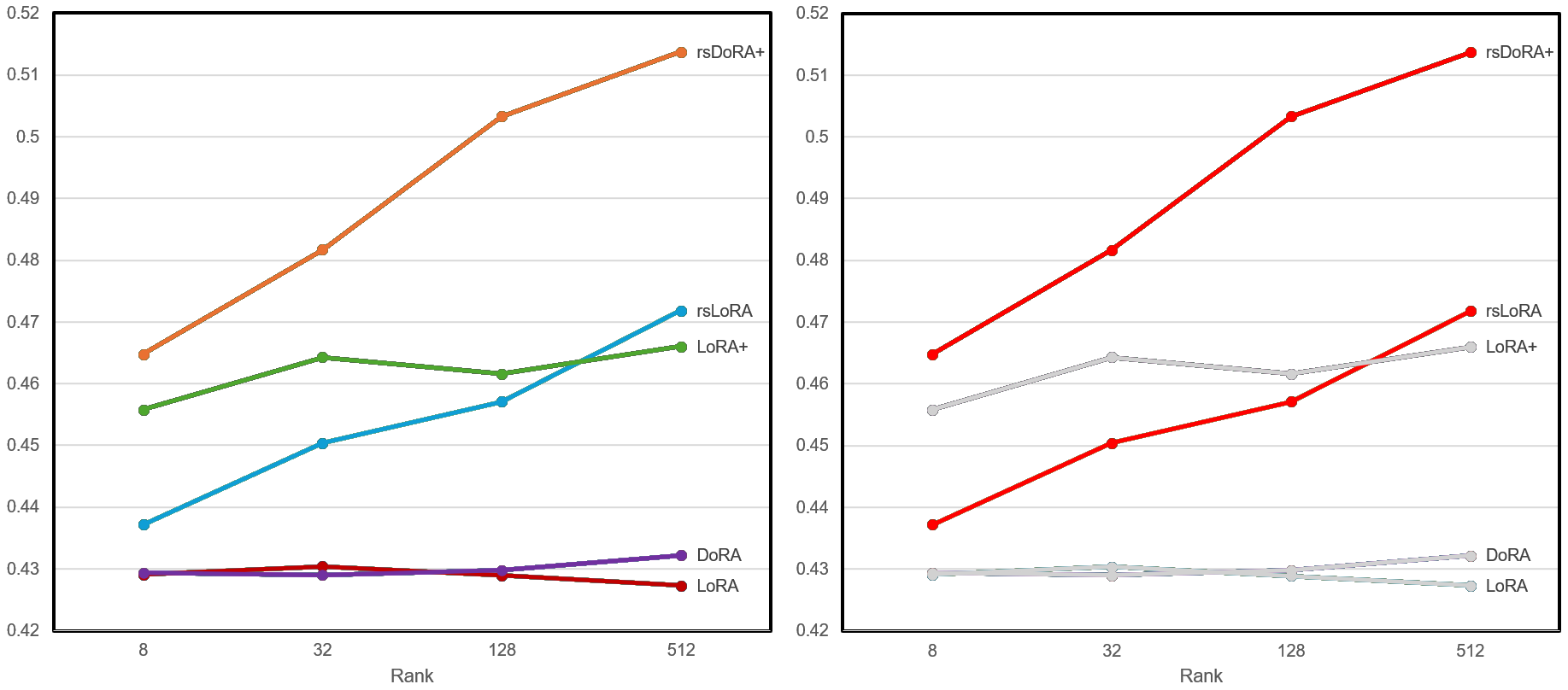}
    \caption{Model Performance at Different Ranks on the Flashcards}
    \label{fig:rankflashcards}
\end{figure}

\begin{figure}[t]
    \centering
    \includegraphics[width=1\linewidth]{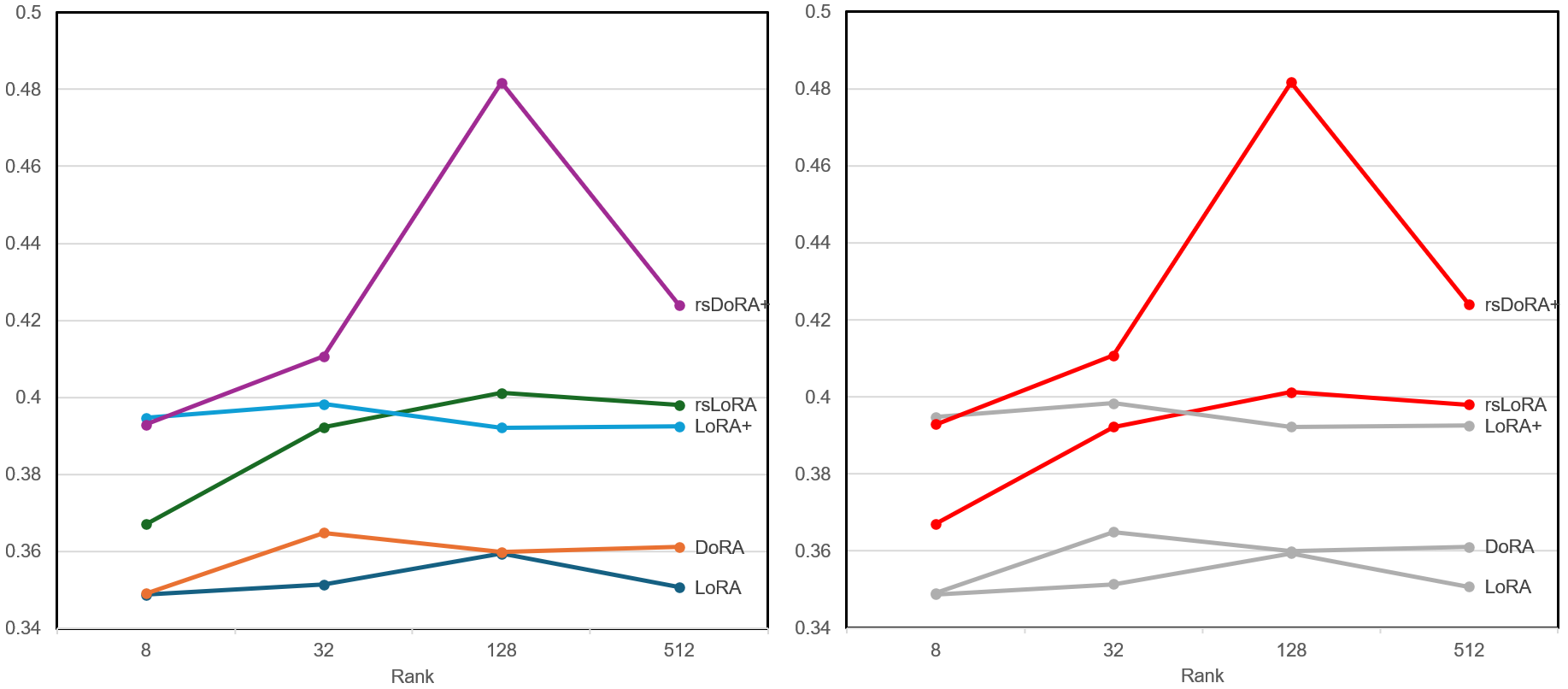}
    \caption{Model Performance at Different Ranks on the MediQA}
    \label{fig:rankMediQA}
\end{figure}

We observed that DoRA and LoRA+ did not perform well in high-rank scenarios and could not achieve effective performance improvements by increasing the rank. By adjusting the adapter scaling factor $\gamma_r$, we made the model's handling of rank more conservative, allowing it to maintain stability at high ranks and achieve performance improvements. Figures \ref{fig:rankMediQA} and \ref{fig:rankflashcards} show that adopting the $\gamma_{rs-L} = \alpha/\sqrt{r}$ strategy allowed rsDoRA+ to inherit the high-rank stability of rsLoRA and amplify the impact of the scaling factor on model performance.

\begin{table}[t]
\centering \caption{Model Performance at Different Ranks on the MediQA}
\begin{tabular}{cccccc}
\hline
 & LoRA & DoRA & rsLoRA & LoRA+ & rsDoRA+ \\
\hline
Rank = 8 & 0.3487 & 0.3491 & 0.3671 & 0.3947 & 0.3929 \\
Rank = 32 & 0.3514 & 0.3649 & 0.3922 & 0.3983 & 0.4107 \\
Rank = 128 & 0.3594 & 0.3599 & 0.4012 & 0.3922 & 0.4817 \\
Rank = 512 & 0.3507 & 0.3611 & 0.3980 & 0.3925 & 0.4240 \\
\hline
\end{tabular}
\end{table}

For the MediQA dataset, the trend of rsDoRA+ and rsLoRA remained consistent. As the rank increased from 8 to 128, the BLEU-4 score significantly improved. Compared to the optimal rank (rank=8) in traditional LoRA methods, the model's performance increased from 0.3929 to 0.4817, an improvement of approximately 10\%. Traditional LoRA and LoRA+ methods exhibited fluctuating performance with increasing rank and even slight declines. When the rank reached 512, the performance of rsDoRA+ and rsLoRA was not as good as at rank=128, with the BLEU-4 score dropping to slightly higher than at rank=32 (0.424). This might be due to the small size of the MediQA dataset (2.2k rows), where a higher rank results in too many parameters without sufficient data support, leading to incomplete training. This was verified in subsequent tests on the Flashcards dataset.

\begin{figure}
    \centering
    \includegraphics[width=1\linewidth]{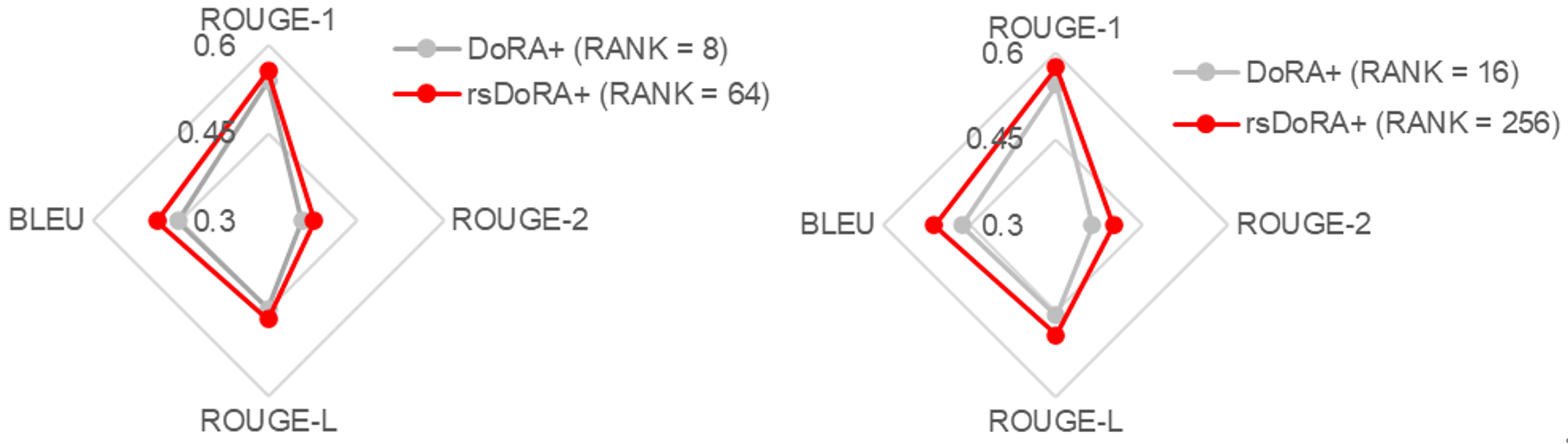}
    \caption{Performance Comparison of rsDoRA+ and DoRA+ with the Same Scaling Factor}
    \label{fig:Same Scaling Factorl}
\end{figure}

For the Anki Flashcards dataset, the BLEU-4 score of rsDoRA+ and rsLoRA gradually increased as the rank rose from 8 to 512. rsDoRA+ improved from 0.4648 at rank=8 to 0.5137 at rank=512, a 5\% performance increase. Given the larger dataset size (34k rows) compared to MediQA, there was enough data to support training at high ranks. Traditional LoRA and LoRA+ methods showed no significant changes at high ranks, with LoRA's performance even deteriorating, as observed with MediQA. These experiments validated the effectiveness of replacing the scaling factor, resulting in significant performance improvements in both datasets. rsDoRA+ inherited the trend observed in rsLoRA. However, the size of the dataset must be considered during high-rank experiments. Insufficient data may lead to negative effects due to the excessive number of parameters.

\begin{table*}[t]
\centering \caption{Growth Rate of rsDoRA+ Compared to DoRA and rsLoRA}
\begin{tabular}{ccccc}
\hline
 & Rank = 8 & Rank = 32 & Rank = 128 & Rank = 512 \\
\hline
DoRA+rsLoRA & 0.4422 & 0.4519 & 0.4773 & 0.4889 \\
rsDoRA+ & 0.4648 & 0.4817 & 0.5033 & 0.5137 \\
Growth Percentage & 5.11\% & 6.59\% & 5.45\% & 5.07\% \\
\hline
\end{tabular}
\label{tab:GrowthRate}
\end{table*}

\subsection{ReRAG Results}
In practical applications, professional databases update rapidly, and the various costs associated with fine-tuning often prevent models from learning the latest knowledge. Additionally, in many fields, including the medical field, information security needs to be considered. The Retrieval-Augmented Generation (RAG) method effectively addresses this issue by enabling the model to obtain real-time accurate information during the generation process. In medical problems, specialized terminology frequently appears. The significance of these keywords is critical in the generation process. However, most retrieval methods often fail to capture these keywords accurately and obtain useful information. Our question rewrite function effectively resolves this issue, ensuring that the retrieved information is more relevant and the generated answers are more accurate.

The ReRAG method's results are shown in Table \ref{tab:Results of ReRAG method comparative experiments}. Mauve (Measuring Automated Unintended Vulnerability and Efficacy) evaluates model generation quality, comparing prediction and real answer distributions. A Mauve score of 1 indicates a perfect match between the prediction's and real answer's distributions, showing fluency, informativeness, and diversity. The baseline for our comparative experiment, rsDoRA+, does not involve retrieval. Traditional RAG, with a vector search component, provides retrieved text for every input. ReRAG - WQE has a similar structure to ReRAG but lacks the question rewrite component.

\begin{table*}[t]
\centering \caption{Results of ReRAG method comparative experiments.}
\begin{tabular}{cccccc}
\hline
  & BLEU & ROUGE-1 & ROUGE-2 & ROUGE-L & Mauve \\
\hline
rsDoRA+ & 0.5137 & 0.5768 & 0.4018 & 0.4919 & 0.7959 \\
RAG & 0.6525 & 0.7112 & 0.5552 & 0.6595 & 0.9236 \\
ReRAG - WQE & 0.7099 & 0.7434 & 0.5974 & 0.6851 & 0.9087 \\
ReRAG & 0.7186 & 0.7508 & 0.6068 & 0.6937 & 0.9502 \\
\hline
\end{tabular}
\label{tab:Results of ReRAG method comparative experiments}
\end{table*}

\begin{figure}[!t]
    \centering
    \includegraphics[width=1\linewidth]{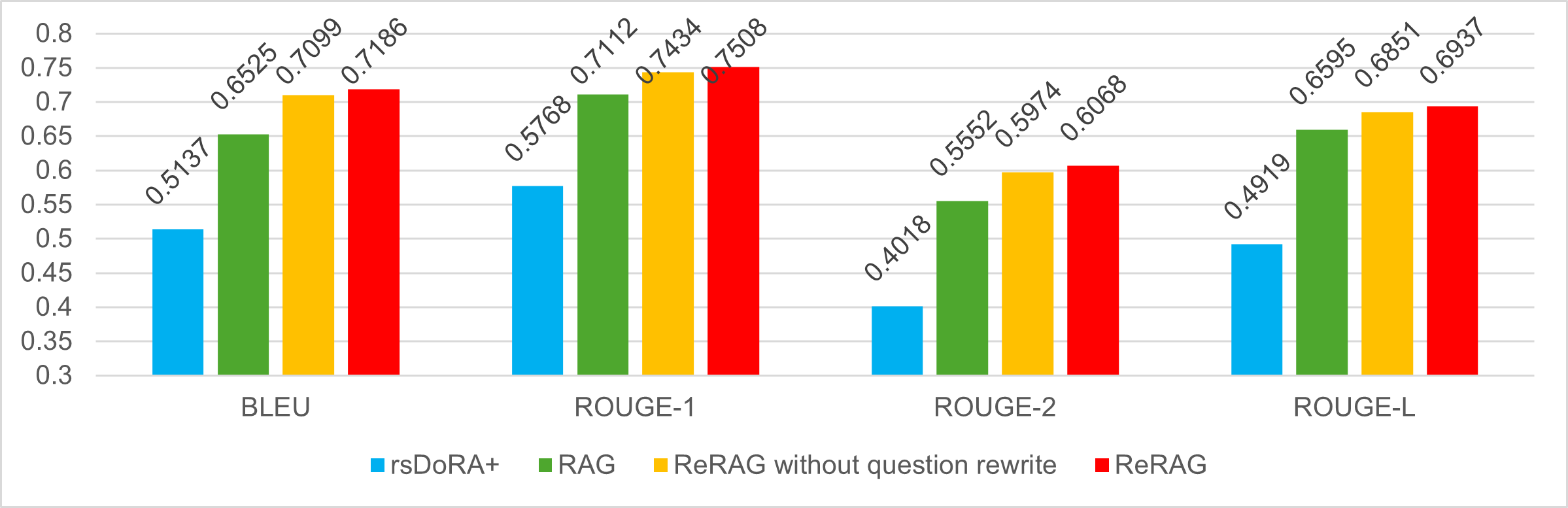}
    \caption{Results of ReRAG method comparative experiments.}
    \label{fig:Results of ReRAG method comparative experiments}
\end{figure}

Figure \ref{fig:Results of ReRAG method comparative experiments} shows that ReRAG outperforms all other methods and the baseline. ReRAG's performance, represented by the red bars, is higher than ReRAG without the question rewrite component, demonstrating the component's effectiveness. Approximately 70\% of questions in the final output included supplementary information, indicating that around 30\% of questions did not require retrieval, highlighting the method's utility.

\subsection{Node Analysis and NEFTune Optimization}
During the experiment, we examined the outputs of each node, especially the IsRel and question rewrite nodes. Some questions required supplementary information, but the IsRel node initially evaluated the retrieved texts as irrelevant. The retriever obtained different texts using key phrases extracted by the question rewrite component, which were then evaluated as useful. For example, the question “What is the potential neurological complication that can arise from tertiary syphilis, and what is the specific area of the nervous system that can be affected by demyelination as a result of this complication?” initially retrieved irrelevant texts. After question rewriting, relevant supplementary information was obtained, proving the component's effectiveness.

Our experimental results show that when $\alpha = 5$, NEFTune performs best. At $\alpha = 5$, our model performs better on long and difficult problems (i.e., the MediQA dataset) compared to normal text length problems (the WikiFlashcard dataset). Medical datasets often contain specialized terms and concepts not commonly seen in general language training. Introducing noise through NEFTune helps the model handle unfamiliar medical terms, reducing overfitting to common language patterns. Adding noise to embedding vectors acts as a random walk in the embedding space, increasing the model's tolerance to input data uncertainty.

Next, we moved to rsDoRA+ experiments. Our results show continuous performance improvement as the rank increases, similar to rsLoRA. Methods without adjusted scaling factors, such as traditional LoRA, perform poorly at high ranks. Traditional LoRA often faces gradient collapse at high ranks due to its scaling factor, leading to slower learning and limited performance improvement. rsDoRA+ solves this problem by adjusting the scaling factor to $\frac{\alpha}{\sqrt{\tau}}$, preventing gradient collapse at high ranks, ensuring stable learning and better performance with the same computational resources.

In practical operations, incorrect retrieved information or added retrieval information for simple questions can lead to inaccurate model generation, a serious issue in the medical field. The ReRAG method successfully solves this problem, ensuring the model obtains more relevant information and simple questions do not receive redundant information. We first fine-tuned the model using rsDoRA+ to optimize performance, then integrated the ReRAG method into the QA system. ReRAG combines the advantages of information retrieval and generation models, retrieving relevant information from a large text database and incorporating it into the answer generation process, enhancing the quality and accuracy of answers.

Medical questions often require extensive background knowledge and precise information. The QA system, integrating rsDoRA+ and ReRAG, provides more accurate answers and delivers results faster when handling complex medical questions. This approach could be a robust solution for medical QA systems, optimizing fine-tuning methods and integrating retrieval-augmented generation technology, showing how combining innovative technologies improves performance in practical applications.

\section{Conclusion}
In this paper, we introduced a comprehensive framework designed to improve large language models (LLMs) for medical question-answering services, with the goal of making healthcare more accessible and reliable. We developed rsDoRA+, an advanced fine-tuning technique that integrates Rank-Stable LoRA, DoRA, and LoRA+ with NEFTune, which helps reduce overfitting and makes learning more stable. This is especially important when processing large amounts of medical data, ensuring the model’s performance remains accurate and consistent. Our experiments showed that this approach significantly enhances LLMs, particularly by carefully managing noise introduction to boost stability. We also introduced ReRAG, a method that combines retrieval-augmented generation with a question rewrite component. This ensures that the information retrieved is relevant and accurate, improving the model’s ability to answer medical questions reliably. When compared to traditional approaches, ReRAG outperformed existing models, offering more precise and contextually appropriate answers. By integrating these techniques, our framework not only tackles technical challenges like overfitting but also enhances the relevance and quality of the information provided. This work underscores the potential of AI-driven healthcare services to offer more accurate and dependable support for healthcare professionals, ultimately contributing to better patient care and more equitable access to medical knowledge for all.

\bibliographystyle{ieeetr}
\bibliography{ref}

\end{document}